\documentclass[letterpaper, 10pt, conference]{ieeeconf}  

\usepackage{url}

\usepackage[utf8]{inputenc}
\usepackage{graphicx}
\graphicspath{{../figures/}}
\usepackage{epsfig}
\usepackage{amsmath}
\usepackage{epstopdf}

\usepackage{amsthm}
\newtheorem{theorem}[subsection]{Theorem}
\usepackage{amssymb}
\usepackage{amsfonts}
\usepackage{cite}
\usepackage[font=small,labelfont=bf]{caption}
\usepackage{comment}
\usepackage[normalem]{ulem}
\usepackage{bbm}
\usepackage{subcaption}
\usepackage{tikz}
\usepackage{tkz-euclide}
\usetikzlibrary{shapes.geometric, arrows}
\usetikzlibrary{positioning,fit,backgrounds}
\usetikzlibrary{arrows, shapes, decorations.markings, calc, patterns, angles, quotes, intersections, plotmarks}

\tikzstyle{controller}=[draw, fill=blue!20, text width=5em, text centered, minimum height=2.5em, rounded corners]
\tikzstyle{annotations} = [above, text width=5em]
\tikzstyle{commblock} = [draw, text width=5em, fill=red!20, minimum height=2.5em, rounded corners, text centered]
\tikzstyle{navblock} = [draw, text width=5em, fill=orange!20, minimum height=2.5em, rounded corners, text centered]
\tikzstyle{cvblock} = [draw, text width=5em, fill=cyan!20, minimum height=2.5em, rounded corners, text centered]
\tikzstyle{compblock} = [draw, text width=5em, fill=magenta!40, minimum height=2.5em, rounded corners, text centered]
\tikzstyle{external} = [draw, text width=5em, fill=blue!40, minimum height=2.5em, rounded corners, text centered]

\usepackage{aircraftshapes}

\usepackage{fontawesome} 

\usepackage{float}
\usepackage[ruled,vlined]{algorithm2e}

\newcommand{\revfin}[1]{{\color{black} #1}}

\usepackage{mathtools}

\usepackage{hyperref}
\usepackage{cleveref}

\title{\LARGE \bf
Probabilistic Consensus on Feature Distribution for Multi-robot Systems with Markovian Exploration Dynamics
}

\author{Aniket Shirsat, Shatadal Mishra, Wenlong Zhang, and  Spring Berman 
\thanks{This work was supported by the Arizona State University Global Security Initiative.}%
\thanks{Aniket Shirsat and Spring Berman are with the School for Engineering of Matter, Transport and Energy, Arizona State University, Tempe, AZ, 85287 USA {\tt\small \{ashirsat, Spring.Berman\}@asu.edu}.}%
\thanks{Shatadal Mishra and Wenlong Zhang are with the Polytechnic School, Arizona State University, Mesa, AZ 85212, USA {\tt\small \{smishr13, Wenlong.Zhang\}@asu.edu}.}
}
\IEEEoverridecommandlockouts 
\begin{document}
\maketitle
\thispagestyle{plain}
\pagestyle{plain}

\begin{abstract}
In this paper, we present a consensus-based decentralized multi-robot approach to reconstruct a discrete distribution of features, modeled as an occupancy grid map, that represent information contained in a bounded planar 2D environment, such as visual cues used for navigation or semantic labels associated with object detection.
The robots explore the environment according to a random walk modeled by a discrete-time discrete-state (DTDS) Markov chain and estimate the feature distribution from their own measurements and the estimates communicated by neighboring robots, using a distributed Chernoff fusion protocol. We prove that under this decentralized fusion protocol, each robot's feature distribution converges to the ground truth distribution in an almost sure sense. We verify this result in numerical simulations that show that the Hellinger distance between the estimated and ground truth feature distributions converges to zero over time for each robot. We also validate our strategy through Software-In-The-Loop (SITL) simulations of quadrotors that search a bounded square grid for a set of visual features distributed on a discretized circle. 
\end{abstract}

\section{Introduction}

Multi-robot systems (MRS) composed of multiple mobile robots have been used for various collective exploration and perception tasks, such as mapping unknown environments \cite{burgard2005coordinated}, disaster response \cite{nagatani2013emergency}, and surveillance and monitoring \cite{mendoncca2016cooperative}. The performance of MRS in such applications is constrained by the capabilities of the payloads that the robots can carry on-board, including the power source, sensor suite, computational resources, and communication devices for transmitting information to other robots and/or a central node. These constraints are particularly restrictive in the case of small aerial robots such as multi-rotors that perform vision-guided tasks \cite{mishra2021tmech}.
Centralized MRS strategies for exploration and mapping, \revfin{such as the next-best-view planning method  in \cite{sukkar2019multi},}
rely on constant communication between all the robots and a central node. Scaling up such strategies with the number of robots requires expanding the communication infrastructure and preventing communication failures of the central node. \revfin{Frontier-based MRS exploration strategies such as  \cite{mahdoui2018cooperative} rely on a dynamic communication topology in which leaders are responsible for coordinating the team.}
\begin{figure}[t]
    \centering
    \includegraphics[width=0.39\textwidth]{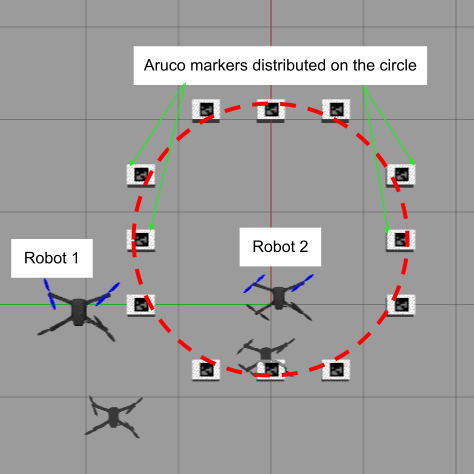}
    \caption{Software-In-The-Loop (SITL) setup in Gazebo using the Rotors \cite{furrer2016rotors} package. Robots 1 and 2 are quadrotors, and the ArUco markers represent a discrete approximation of a continuous circular feature (the red dotted line).} \vspace{-5mm}
    \label{fig:sitl_setup}
\end{figure}




Decentralized MRS exploration and mapping strategies that employ only local communication alleviate these drawbacks and are designed to work robustly under inter-robot communication bandwidth constraints \cite{howard2006experiments} and disruptions to communication links by environmental effects \cite{husain2013mapping}. Many decentralized MRS estimation strategies are designed to achieve {\it consensus} among the robots on a particular variable or property of interest through local inter-robot communication. For example, distributed consensus-based approaches have been designed for spacecraft attitude estimation \cite{li2019fully} and space debris tracking \cite{wei2017consensus}. Consensus behaviors also arise in social networks \cite{xia2015structural} when users reach an agreement on a shared opinion in a distributed fashion.
Consensus strategies have been developed for MRS communication networks that are static or dynamic, and that can be represented as directed or undirected graphs \cite{ren2004consensus}, as well as random networks \cite{mesbahi2010graph} and networks with communication delays \cite{olfati2004consensus}. However, few works address consensus problems for MRS that follow random mobility models, often used in MRS exploration strategies as in, e.g., \cite{kegeleirs2019random}, whose communication networks exhibit Markovian switching dynamics as a result. Random exploration strategies have certain advantages for MRS: they do not require centralized motion planning, localization, or communication, and they can be modified to produce more focused or more dispersed coverage.

In our previous work \cite{shirsat2020multirobot}, we developed a probabilistic consensus-based strategy for target search by MRS with a discrete-time discrete-state (DTDS) Markov motion model and local sensing and communication. Using this strategy, the robots are guaranteed to achieve consensus almost surely on the presence of a static feature of interest, without any requirements on the connectivity of the robots' communication network. We extended this approach in \cite{shirsat2021decentralized} to an MRS strategy for tracking multiple static features by formulating the tracking procedure as a renewal-reward process on the underlying Markov chain.
The robots reach a consensus on the number of features and their locations in a decentralized manner using a Gaussian Mixture approximation of the Probability Hypothesis Density (PHD) filter.

In this paper, we generalize the consensus objective of our probabilistic multi-robot search strategy to agreement on a discrete distribution of static features, modeled as an occupancy grid map, 
using results on \textit{opinion pools}\cite{bailey2012conservative}.
We consider a group of robots that move according to a DTDS Markov chain on a finite 2D spatial grid, as shown in \Cref{fig:MotionStrategy_Visual}, and that can detect features using their on-board sensors. The proposed strategy is distributed and asynchronous, and it preserves the required communication bandwidth by relying only on local inter-robot communication.
The main contributions of the paper are as follows:  
\begin{enumerate} 
\item  \revfin{We present a decentralized, stochastic multi-robot exploration and mapping strategy in which the robots use a consensus protocol, without communication connectivity requirements, to arrive at a common reconstruction of a feature distribution on a 2D grid.} Specifically, given a group of robots with a  DTDS Markov motion model and local sensing and communication, we prove that if the robots update their estimates of the feature distribution with those of other robots according to a distributed Chernoff fusion protocol, then they will reach consensus almost surely on the ground truth 
distribution. 
This extends the result in \cite{battistelli2014kullback} on opinion consensus over fixed, strongly connected networks to networks with Markovian switching dynamics.
\item We validate our theoretical results in numerical simulations that illustrate the pathwise convergence to zero of the Hellinger distance between each robot's estimate of the feature distribution and the ground truth distribution.
    We also validate our approach in Software-In-The-Loop (SITL) simulations of quadrotors, performed in Gazebo using the Robot Operating System (ROS) with the PX4 autopilot.
\end{enumerate}


The remainder of the paper is organized as follows. We present our probabilistic  exploration strategy and information fusion protocol in \Cref{sec:ProbStat}. We describe some relevant properties of DTDS Markov chains in \Cref{sec:MarkoExplorDyn}, and we derive the main result that guarantees the convergence of each robot's feature distribution to the ground truth distribution in \Cref{sec:ConsensusMarkov}. We present the results of our numerical simulations in \Cref{sec:MatSims} and our SITL simulations in \Cref{sec:Expts}. \Cref{sec:Conc} concludes the paper and suggests directions for future work. \revfin{A video overview of the paper is provided at \url{https://youtu.be/-Z4-DZrHwSM}.}

\section{Exploration and Information Fusion Strategy}\label{sec:ProbStat}
Consider a bounded square environment $\mathcal{B} \subset \mathbb{R}^2$ with sides of length $B$. We discretize $\mathcal{B}$ into a square grid of nodes spaced at a distance $\delta$ apart. The set of nodes is denoted by $\mathcal{S} \subset \mathbb{Z}_+$, and we define $S = |\mathcal{S}|$. A set of $N$ robots, $\mathcal{A} = \{1,2,\ldots,N\}$, each modeled as a point mass, explore the environment by performing a random walk on the grid. We assume that there are no obstacles in the environment that impede the robots' motion. Let $\mathcal{G}_{s} = (\mathcal{V}_{s}, \mathcal{E}_{s})$ be an undirected graph associated with this finite spatial grid, where $\mathcal{V}_{s} =\mathcal{S}$ is the set of nodes and $\mathcal{E}_{s}$ is the set of edges $(i,j)$. The edges  signify pairs of nodes $i,j \in \mathcal{V}_s$, called {\it neighboring nodes}, between which robots can travel. We assume that the robots can localize on $\mathcal{G}_s$.

Let $Y^a_k \in \mathcal{S}$ be a random variable that defines the node that robot $a \in \mathcal{A}$ occupies at discrete time $k$.
Robot $a$ moves from its current node $i$ to a neighboring node $j$ at the next time step with a transition probability $p_{ij} \in [0,1]$. We define $\mathbf{P} \in \mathbb{R}^{S \times S}$ as the \textit{state transition matrix} consisting of elements $p_{ij}$ at row $i$ and column $j$. Let $\pi_{k} \in \mathbb{R}^{1 \times S}$ denote the probability mass function (PMF) of $Y^a_k$ for each robot $a$, or alternatively, the distribution of the robot population over the grid at time $k$. 
This distribution evolves over time according to a DTDS Markov chain model of order one:
\begin{equation}
    \pi_{k+1} = \pi_{k} \mathbf{P}.
    \label{eqn:MarkovChain}
\end{equation}
We assume that each robot can exchange information with other robots that are within a communication radius $r_{comm} < 0.5\delta$. Let $\mathcal{G}_c[k] =  (\mathcal{V}_{c},\mathcal{E}_{c}[k])$ be an  undirected graph in which  $\mathcal{V}_{c} = \mathcal{A}$, the set of robots, and $\mathcal{E}_{c}[k]$ is the set of all pairs of robots $(a,b) \in \mathcal{A} \times \mathcal{A}$ that can communicate with each other at time $k$.  Let $\mathbf{M}[k] \in \mathbb{R}^{N \times N}$ be the {\it adjacency matrix} with elements $m_{ab}[k] = 1$ if $(a,b) \in \mathcal{E}_c[k]$  and $m_{ab}[k] = 0$ otherwise. For each robot $a \in \mathcal{A}$, we define the set of {\it neighbors} of robot $a$ at time $k$ as $\mathcal{N}^{a}_{k} \triangleq \{b \in \mathcal{A}: (a,b) \in \mathcal{E}_{c}[k] \}$.
 
A set of discrete features is distributed over the grid at nodes in the set $\mathcal{B}^{r} \subseteq \mathcal{S}$. The robots know a priori that these features are present in the environment, but do not know their distribution. We assume that when a robot is located at a node in $\mathcal{B}^{r}$, it can detect the presence of a feature at that node using its on-board sensors. 
Each node in the grid is associated with  a binary occupancy value, defined as $\bar{l} \in (0.5, 1)$ if the robot detects a feature at that node and $1-\bar{l}$ if it does not. Setting  $\bar{l} \in (0.5,1)$ helps produce sharp reconstructions of the features: as the value of $\bar{l}$ increases, the distinction between occupied and unoccupied nodes becomes clearer.

We define the {\it occupancy vector} for robot $a$ at time $k$ as $\bar{\theta}^{a}_{k} = [\theta^{a}_{k}(1)~ \ldots~ \theta^{a}_{k}(S)] \in \mathbb{R}^{1 \times S}$, where 
\begin{equation}
    \theta^{a}_{k}(s) =
    \begin{cases}
    \bar{l}, & s  \in \mathcal{B}^{r} ~(occupied) \\
    1-\bar{l}, & o.w ~ (unoccupied)
    \end{cases} 
    \label{eqn:OccupancyGridFunction}
\end{equation}
The occupancy vector for each robot indicates its estimate of the nodes that are occupied by features. The feature PMF, or \textit{occupancy distribution}, estimated by robot $a$ at time $k$ from its own sensor measurements \revfin{is defined as:}
\begin{equation}
    f^{a}_{k}(s) =\frac{\theta^{a}_{k}(s)}{\sum_{i \in \mathcal{S}} \theta^{a}_{k}(i)}
    \label{eqn:feautePMF}
\end{equation}
\revfin{Here, $f^{a}_{k}(s)$ also represents the \textit{opinion} \cite{degroot1974reaching} 
of robot $a$ at time $k$ for the occupancy distribution.}
We denote $\theta^{ref} \in \mathbb{R}^{1 \times S}$  
as the 
\textit{reference occupancy vector} that is being estimated by all the robots, defined as follows:    
\begin{equation} 
    \theta^{ref}(s) =
    \begin{cases}
    \bar{l}, & s  \in \mathcal{B}^{r} \\
    1-\bar{l}, & o.w
    \end{cases} 
    \label{eqn:RefOccupancyFunction}
\end{equation}
Since the robots do not know the occupancy distribution a priori, we specify that they all initially consider the grid to be unoccupied, i.e.,
\begin{equation} 
     f^{a}_{0}(s) = \frac{1-\bar{l}}{S(1 - \bar{l})}, ~~~a \in \mathcal{A}, ~\forall s \in \mathcal{S}
\end{equation}
This is defined as the 
\textit{nominal distribution} for all the robots, denoted by $f^{nom}(s) = f^{(\cdot)}_{0}(s)$. We also denote a vector $\theta^{nom}  \in \mathbb{R}^{1 \times S}$ as the {\it nominal occupancy vector} for all robots, which 
represents all nodes as unoccupied (i.e., $\theta^{nom}(s) = \theta^{(\cdot)}_{0}(s) = 1 -\bar{l}, ~\forall s \in \mathcal{S}$). We define $f^{ref}(s)$ as the 
\textit{reference PMF}, the ground truth feature distribution, 
 corresponding to $\theta^{ref}(s)$.
We define a fusion weight $\omega^{(a,b)}_k$ as the following \textit{Metropolis weight} \cite{calafiore2009distributed}: 
\begin{equation}
    \omega^{(a,b)}_k =  
    \begin{cases}
    \frac{1}{1+|\mathcal{N}_{k}^{b}|}, & b \in \mathcal{N}_{k}^{a}\setminus\{a\} \\
    1 - \sum_{b \in \mathcal{N}_{k}^{a}\setminus\{a\}} \omega^{(a,b)}_k, & a=b, ~a \in \mathcal{A} \\
    0, & o.w
    \end{cases}
    \label{eqn:MetroWeights}
\end{equation}
Note that $\omega^{(a,b)}_{k} \geq 0$ and $\sum_{b \in \mathcal{N}_{k}^a} \omega^{(a,b)}_k = 1$. We then define the {\it consensus} or {\it opinion weighting matrix} $\Omega_k \in \mathbb{R}^{N \times N}$ 
at time $k$, which consists of elements $\omega^{(a,b)}_k$ at row $a$ and column $b$.

Given the robot exploration dynamics in \Cref{eqn:MarkovChain}, each robot $a$ updates its opinion $f_k^a(s)$, computed from \Cref{eqn:feautePMF}, to the following PMF $f^{cher}_{k+1}(s)$ at the next time step, which it computes from the opinions of other robots within its communication range according to the {\it Chernoff fusion rule} 
\cite{farrell2009generalized}: 
\begin{equation} 
    f^{cher}_{k+1}(s) = \frac {\prod_{b \in \mathcal{N}_{k}^{a} \cup \{a\}} [f^{b}_{k}(s) ]^{\omega^{(a,b)}_k}}{\sum_{s \in \mathcal{S}} \prod_{b \in \mathcal{N}_{k}^{a} \cup \{a\}} [f^{b}_{k}(s) ]^{\omega^{(a,b)}_k}}, ~~ 
    a \in \mathcal{A}
    \label{eqn:Opinion_Update_eqn}
\end{equation}
Applying Theorem 1 from \cite{battistelli2014kullback}, we can say that $f^{cher}_{k+1}(s)$ is the local neighbor fused feature PMF at time $k+1$ of the all robots $a \in \mathcal{A}$ that are in $\mathcal{N}^{a}_{k}$. 
When only two robots $a,b \in \mathcal{A}$ are within communication range, this update rule becomes:
\begin{equation}
    f^{cher}_{k+1}(s) = \frac{[f^a_k(s)]^{\omega} [f^b_k(s)]^{1- \omega}}{\sum_{s \in \mathcal{S}} [f^a_k(s)]^{\omega} [f^b_k(s)]^{1- \omega}},
    \label{eqn:Chernoff_two_agents}
\end{equation}
where the Metropolis weight $\omega \equiv \omega^{(a,b)}_k$
is defined in \Cref{eqn:MetroWeights}. 
Then each robot $a$ compares $f^{cher}_{k+1}(s)$ with $f^{nom}(s)$ to generate a new fused occupancy vector as follows:   
\begin{equation}
\theta^{cher}_{k+1}(s) =
\begin{cases}
\bar{l} & f^{cher}_{k+1}(s) > f^{nom}(s), ~ s \in \mathcal{S}\\
1 - \bar{l} & o.w \\
\end{cases}
\label{eqn:FusedOccupancyVec}
\end{equation}
Then, each robot $a$ generates a new occupancy vector $\theta^{a}_{k+1}(s)$ by comparing both its occupancy vector at the previous time step, $\theta^{a}_{k}(s)$, and the fused occupancy vector $\theta^{cher}_{k+1}(s)$, to 
$\theta^{nom}(s)$:
\begin{equation}
    \theta^{a}_{k+1}(s) =
    \begin{cases}
    \bar{l} & \theta^{a}_{k}(s) > \theta^{nom}(s) ~or\\
    & ~ \theta^{cher}_{k+1}(s) > \theta^{nom}(s), ~s \in \mathcal{S} \\
    \theta_{k}^{a}(s) & o.w
    \end{cases}
    \label{eqn:NextOccupancyVec}
\end{equation}

To quantify the convergence of each robot's feature distribution to the reference distribution, we use the \textit{Hellinger metric}, which measures the similarity between two PMFs. 
The Hellinger distance between the feature PMF $f^{a}_k(s)$ of a robot $a \in \mathcal{A}$ at time $k$ and the reference PMF $f^{ref}(s)$ is given by 
\begin{equation}
    \mathbf{D}_{H}(f^a_k(s),f^{ref}(s)) = \sqrt{1 - \rho(f^a_k(s),f^{ref}(s))}~,
    \label{eqn:Hell_dist}
\end{equation}
where $\rho(f^{a}_{k}(s),f^{ref}(s))$ is the \textit{Bhattacharya coefficient}, defined as: 
\begin{equation}
    \rho(f^a_k(s),f^{ref}(s)) = \sum_{s \in \mathcal{S}} \sqrt{f^a_k(s) \cdot f^{ref}(s)}~.
    \label{eqn:Bhatta_coeff}
\end{equation}

This distance takes values in $[0,1]$. 
We define the vector $\mathbf{D}_{H} \in \mathbb{R}^{N \times 1}_{\geq 0}$ with each entry $a \in \mathcal{A}$ given by $\mathbf{D}_{H}(f^a_k(s),f^{ref}(s))$.
\begin{algorithm}[t]
\SetAlgoLined
\DontPrintSemicolon
\SetNoFillComment
\SetKwInOut{Input}{input}
\SetKwInOut{Output}{output}
\SetKwInput{Initialize}{initialization}

\Input{$f^{a}_k, ~ f^{b}_k, ~ Y^{a}_{k}, ~ Y^{b}_{k}, ~k, ~T, ~\epsilon$}
\Output{$f^{a}_{k+1}$}
\If{$1 < k \leq T$}
{
\eIf{$Y^{a}_{k} = Y^{b}_{k}$}
{
$\omega =0.5$\\ 
}
{ 
$\omega = 1.0$  
}
$c =  \sum_{s \in \mathcal{S}}[f^{a}_{k}(s)]^{\omega} \cdot [f^{b}_{k}(s)]^{1 - \omega}$ \\
$logf_{cher}= \omega \log{(f^{a}_{k})} + (1-\omega) \log{(f^{b}_{k})} - \log{(c)}$\\
$f^{a}_{k+1} = f^{cher}_{k+1} =\exp(logf_{cher})$\\ 
}
\caption{Distributed Chernoff fusion protocol for robots $a, b \in \mathcal{A}$ computed by robot $a$ at time $k$} 
\label{algo:Chernoff_Fusion}
\end{algorithm}
The pseudo code in Algorithm \ref{algo:Chernoff_Fusion} implements this fusion strategy for two robots $a$ and $b$. In this algorithm, the normalizing constant $c$ is the denominator of \Cref{eqn:Chernoff_two_agents}. 

\begin{figure}[t]
    \centering
        \begin{tikzpicture}[scale=0.6]
            \draw [step=1cm, thin, gray!50] (0,0) grid (10,10);
            \filldraw[color=magenta] (3-0.2,5-0.2) rectangle (3+0.2,5+0.2);
            \filldraw[color=magenta] (7-0.2,5-0.2) rectangle (7+0.2,5+0.2);
            \filldraw[color=magenta] (5-0.2,7-0.2) rectangle (5+0.2,7+0.2);
            \filldraw[color=magenta] (5-0.2,3-0.2) rectangle (5+0.2,3+0.2);
            \filldraw[color=magenta] (3-0.2,4-0.2) rectangle (3+0.2,4+0.2);
            \filldraw[color=magenta] (7-0.2,4-0.2) rectangle (7+0.2,4+0.2);
            \filldraw[color=magenta] (3-0.2,6-0.2) rectangle (3+0.2,6+0.2);
            \filldraw[color=magenta] (7-0.2,6-0.2) rectangle (7+0.2,6+0.2);
            \filldraw[color=magenta] (4-0.2,3-0.2) rectangle (4+0.2,3+0.2);
            \filldraw[color=magenta] (6-0.2,3-0.2) rectangle (6+0.2,3+0.2);
            \filldraw[color=magenta] (4-0.2,7-0.2) rectangle (4+0.2,7+0.2);
            \filldraw[color=magenta] (6-0.2,7-0.2) rectangle (6+0.2,7+0.2);
            \draw[color=magenta] (5,5) circle (2);
            \filldraw[color=black]  (5,5) circle (0.1);
            \node[right] (i) at (4.8,4.5) {$m=Y^1_k=Y^2_k$};
	        \node [quadcopter top,fill=white,draw=cyan,minimum width=0.5cm, rotate=45] at (3.0,1.0) {};
	        \node[fill=cyan,regular polygon, regular polygon sides=3,inner sep=2.5pt] at (2.0,1) {};
	        \node[above] at (1.5,0.0) {$Y^1_0=i$}; 
            \node [quadcopter top,fill=white,draw=orange,minimum width=0.5cm, rotate=45] at (5,9) {};
            \node[fill=orange,regular polygon, regular polygon sides=3,inner sep=2.5pt] at (5,8) {}; 
        	\node[above] at (4.2,7.1) {$Y^2_0 = j$};
        	\draw[->,color=cyan, thick,dashed] (2,1) -- (1,1);
        	\draw[->,color=cyan, thick,dashed] (1,1) -- (1,2);
        	\draw[->,color=cyan, thick,dashed] (1,2) -- (2,2); 
        	\draw[->,color=cyan, thick,dashed] (2,2) -- (2,3);
        	\draw[->,color=cyan, thick,dashed] (2,3) -- (2,4);
        	\draw[->,color=cyan, thick,dashed] (2,4) -- (3,4);
        	\draw[->,color=cyan, thick,dashed] (3,4) -- (4,4);
        	\draw[->,color=cyan, thick,dashed] (4,4) -- (4,5);
        	\draw[->,color=cyan, thick,dashed] (4,5) -- (5,5);
        	\draw[->,color=orange, thick,dashed] (5,8) -- (6,8);
        	\draw[->,color=orange, thick,dashed] (6,8) -- (7,8);
        	\draw[->,color=orange, thick,dashed] (7,8) -- (8,8);
        	\draw[->,color=orange, thick,dashed] (8,8) -- (8,7);
        	\draw[->,color=orange, thick,dashed] (8,7) -- (7,7);
        	\draw[->,color=orange, thick,dashed] (7,7) -- (7,6);
        	\draw[->,color=orange, thick,dashed] (7,6) -- (6,6);
        	\draw[->,color=orange, thick,dashed] (6,6) -- (6,5);
        	\draw[->,color=orange, thick,dashed] (6,5) -- (5,5); 
        	\node [quadcopter top,fill=white,draw=black, minimum width=1.0cm, rotate=45] at (8,2) {};
            \draw [->, color=black, thick] (8,3) -- node[right] {up} (8,4);
            \draw [->, color=black, thick] (7,2) -- node[below]{left} (6,2);
            \draw [->, color=black, thick] (8,1) -- node[right]{down} (8,0);
            \draw [->, color=black, thick] (9,2) -- node[below]{right} (10,2);
        \end{tikzpicture}
        \caption{Illustration of our multi-robot exploration
        strategy, showing  sample paths for two quadrotors (orange and  blue) on a square grid. The quadrotors search the environment for a set of static features (the magenta squares representing a discretized  circle) as they perform a random walk on the grid.} 
        \label{fig:MotionStrategy_Visual}
\end{figure}
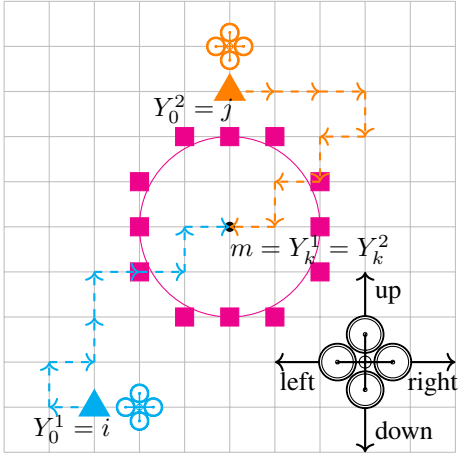

 \Cref{fig:MotionStrategy_Visual} illustrates the proposed strategy for a scenario with two quadrotors. The quadrotors start at the spatial grid nodes indexed by $i$ and $j$ and move on the grid according to the DTDS Markov chain dynamics in \eqref{eqn:MarkovChain}. The figure shows sample paths of the quadrotors. The orange quadrotor detects the feature, indicated by a magenta square, when it moves to a node in the set $\mathcal{B}^r$ (at these nodes, the feature is within the quadrotor's sensing range).  
The quadrotors meet at grid node $m$ after $k=9$ time steps and fuse occupancy distributions according to \Cref{eqn:Chernoff_two_agents}. They continue to random-walk on the grid and update $f_k^a(s)$ until a specified final time $T$. \revfin{
We chose $T$ empirically based on feature PMF convergence times from numerical simulation studies, shown in \Cref{fig:c_vs_nc_comp}, and on flight time constraints of off-the-shelf quadrotors.} 
 
\section{Properties of DTDS Markov Chains} \label{sec:MarkoExplorDyn}

The Markov chain in \Cref{eqn:MarkovChain} is characterized in terms of the time-invariant state transition matrix $\mathbf{P}$, which is defined by the state space of the spatial grid representing the discretized environment. Hence, the Markov chain is $\textit{time-homogeneous}$, which implies that $Pr(Y^a_{k+1} = j ~|~ Y^a_k = i)$ is the same for all robots at all times $k$. The entries of $\mathbf{P}$, which are the state transition probabilities, can therefore be defined as
\begin{equation}
    p_{ij}= Pr(Y_{k+1}^{a} = j ~|~ Y_{k}^{a} = i),~ \forall i,j \in \mathcal{S}, ~k \in \mathbb{Z}_+, ~\forall a \in \mathcal{A}.
    \label{eqn:TransitionProbabilityMatHomogenousMC}
\end{equation}
Since each robot chooses its next node from a uniform distribution, these entries can be computed as 
\begin{equation}
         p_{ij} =\begin{cases} 
         \frac{1}{d_{i}+1}, & (i,j) \in \mathcal{E}_{s}, \\
          0, & o.w, 
    \end{cases}
    \label{eqn:TransitionMat_Elements}
\end{equation}
where $d_{i}$ is the degree of the node $i \in \mathcal{S}$, defined as $d_i = 2$ if  $i$ is a corner node of the spatial grid, $d_i = 3$ if it is on an edge between two corner nodes, and $d_i = 4$ otherwise. 
Since each entry $p_{ij} \geq 0$, we use the notation $\mathbf{P} \geq 0$. We see that $\mathbf{P}^{m} \geq 0$ for $m \geq 1$, and therefore $\mathbf{P}$ is a \textit{non-negative matrix.} From Theorem 5 in \cite{grimmett2001probability}, we can conclude that $\mathbf{P}$ is a stochastic matrix. 
If $\pi$ is a stationary distribution of Markov chain \eqref{eqn:MarkovChain}, 
then $\forall k \in \mathbb{Z}_+$, 
\begin{equation}
     \pi \mathbf{P}^{k} = \pi.
    \label{eqn:StationaryDist}
\end{equation}
From the construction of the Markov chain, 
each robot has a positive probability of moving from any node $i \in \mathcal{S}$ to any other node $j \in \mathcal{S}$ of the spatial grid in a finite number of time steps.
As a result, the Markov chain is {\it irreducible}, and therefore $\mathbf{P}$ is an irreducible matrix. By Lemma 8.4.4 (Perron-Frobenius) in \cite{horn1990matrix}, 
there exists a real unique positive left eigenvector of $\mathbf{P}$ and since $\mathbf{P}$ is a stochastic matrix, its spectral radius $\rho(\mathbf{P})$ is 1. 
Therefore, this left eigenvector is the stationary distribution of the corresponding Markov chain. Since we have shown that the Markov chain is irreducible and has a stationary distribution $\pi$, which satisfies $\pi \mathbf{P} = \pi$, we can conclude from Theorem 21.12 in \cite{levin2017markov} that the Markov chain is {\it positive recurrent}. Thus, all states in the Markov chain are positive recurrent, which implies that each robot will keep visiting every state on the finite spatial grid infinitely often.
\section{Consensus on the Feature Distribution}\label{sec:ConsensusMarkov}
By Theorem 1 of \cite{battistelli2014kullback}, \Cref{eqn:Chernoff_two_agents} achieves opinion consensus over a graph with a fixed and strongly connected topology. We extend this result to graphs with topologies that evolve according to the switching dynamics on the composite Markov chain described in this section. 
We demonstrate that under our opinion fusion scheme, all the robots will reach a consensus on the feature distribution. 

The dynamics of all robots' movements on the spatial grid can be modeled by a composite Markov chain with states 
$\mathbf{Y}_k = (Y^1_k, Y^2_k, \ldots, Y^N_k) \in \mathcal{M}$, where $\mathcal{M} = \mathcal{S}^{\mathcal{A}}$. Note that $S = |\mathcal{S}|$ and $|\mathcal{M}| = S^N$.
We define an undirected graph $\hat{\mathcal{G}} = (\hat{\mathcal{V}},\hat{\mathcal{E}})$ that is associated with the composite Markov chain. 
The vertex set $\hat{\mathcal{V}}$ is the set of all possible realizations $\hat{\imath} \in \mathcal{M}$ of $\mathbf{Y}_k$. The notation $\hat{\imath}(a)$ represents the $a^{th}$ entry of $\hat{\imath}$, which is the 
node $i \in \mathcal{S}$ occupied by robot $a$.
We define the edge set $\hat{\mathcal{E}}$ 
as follows: $(\hat{\imath},\hat{\jmath}) \in \hat{\mathcal{E}}$ if and only if $(\hat{\imath}(a),\hat{\jmath}(a)) \in \mathcal{E}_s$ for all robots $a \in \mathcal{N}$. Let $\mathbf{Q} \in \mathbb{R}^{ | \mathcal{M}| \times |\mathcal{M}|}$ be the state transition matrix associated with the composite Markov chain.
The elements of $\mathbf{Q}$, denoted by $q_{\hat{\imath} \hat{\jmath}}$, are computed from the transition probabilities defined in \Cref{eqn:TransitionMat_Elements} as follows: 
\begin{equation}
    q_{\hat{\imath} \hat{\jmath}} = \prod_{a=1}^{N} p_{\hat{\imath}(a) \hat{\jmath}(a)}, ~~~~ \forall \hat{\imath}, \hat{\jmath} \in \mathcal{M}. 
    \label{eqn:TransitionRatesProductState}
\end{equation}
Here, $q_{\hat{\imath} \hat{\jmath}}$ is the probability that in the next time step, each robot $a$ will move from 
node $\hat{\imath}(a)$ to node $\hat{\jmath}(a)$.

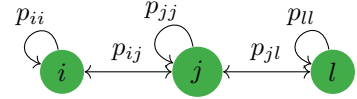
\begin{figure}[t]
    \centering
    \begin{tikzpicture}[scale=0.6,auto=center]
        \node[style={circle,fill=yellow!50!orange!25!green}] (a1) at (0,0) {$i$};
        \node[style={circle,fill=yellow!50!orange!25!green}] (a2) at (3,0) {$j$};
        \node[style={circle,fill=yellow!50!orange!25!green}] (a3) at (6,0) {$l$};
        \node[] (a4) at (5,0){};
        \draw[<->,style={draw=black}] (a1) -- node[above]{$p_{i j}$} (a2);
        \draw[<->,style={draw=black}] (a2) -- node[above]{$p_{j l}$} (a3);
        \draw[->,style={draw=black}] (a1) to [in=160, out=100,looseness=5] node[above] {$p_{i i}$} (a1);
        \draw[->,style={draw=black}] (a2) to [in=160, out=100,looseness=5] node[above] {$p_{j j}$} (a2);
        \draw[->,style={draw=black}] (a3) to [in=160, out=100,looseness=5] node[above] {$p_{l l}$} (a3);
    \end{tikzpicture}
    \caption{A graph $\mathcal{G}_s =  (\mathcal{V}_s,\mathcal{E}_s)$  defined on the set of spatial nodes  $\mathcal{V}_s = \{i,j,l\}$. The arrows signify undirected edges between pairs of distinct nodes and self-edges. The edge set of the graph is $\mathcal{E}_s = \{(i,i), (j,j), (l,l), (i,j), (j,l)\}$.}
    \label{fig:MarkovianTransition} 
\end{figure} 
\begin{figure}[t]
    \centering
    \begin{tikzpicture}[scale=.4,auto=center]
        \node  [style={circle,fill=blue!30!white}] (a1i1a2i1) at (0,0) {$(i,i)$};
        \node (ghatv1) at (0,2.0) {$\hat{i}$};
        \node [style={circle,fill=blue!30!white}] (a1i1a2i2) at (5,0) {$(i,j)$};
        \node (ghatv2) at (5,2.0) {$\hat{j}$};
        \node  [style={circle,fill=blue!30!white}] (a1i1a2i3) at (10,0) {$(i,l)$};
        \node (ghatv3) at (10,2.0) {$\hat{l}$}; 
        \node (ghatend) at (13,0){}; 
        \draw[<->,style={draw=black}] (a1i1a2i1) -- node[above,midway]{$q_{\hat{i},\hat{j}}$} (a1i1a2i2);
        \draw[<->,style={draw=black}] (a1i1a2i2) -- node[above,midway]{$q_{\hat{j},\hat{l}}$} (a1i1a2i3);
        \draw[style={draw=black, dashed}] (a1i1a2i3) -- (ghatend);
        \draw[<->,style={draw=black}] (a1i1a2i1) to [loop, in=250, out=210,looseness=3] node[below] {$q_{\hat{i},\hat{i}}$} (a1i1a2i1);
        \draw[<->,style={draw=black}] (a1i1a2i2) to [loop, in=250, out=210,looseness=3]
        node[below] {$q_{\hat{j},\hat{j}}$} (a1i1a2i2);
        \draw[<->,style={draw=black}] (a1i1a2i3) to [loop, in=250, out=210,looseness=3] 
        node[below] {$q_{\hat{l},\hat{l}}$} (a1i1a2i3);
    \end{tikzpicture}
    \caption{A subset of the composite graph $\mathcal{\hat{G}}=(\mathcal{\hat{V}},\mathcal{\hat{E}})$ for 2 robots that 
    move on the graph $\mathcal{G}_s$ shown in \Cref{fig:MarkovianTransition}.}
    \label{fig:Q_Mat_elem_Vis}
\end{figure}
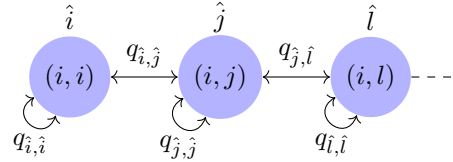

For example, consider a set of two robots, $\mathcal{N}$ = $\{1,2\}$, that move on the graph $\mathcal{G}_{s}$ 
in \Cref{fig:MarkovianTransition}. In the next time step, the robots can stay at their current node 
or travel between nodes $i$ and $j$ and between nodes $j$ and $l$, but they cannot travel between nodes $i$ and $l$.  
\Cref{fig:Q_Mat_elem_Vis} shows a subset of the resulting composite graph $\mathcal{\hat{G}}$,whose entire set of nodes is $\hat{\mathcal{V}}$ = $\{(i,i), (i,j), (i,l), (j,i), (j,j), (j,l), (l,i), (l,j), (l,l)\}$. Each node in $\hat{\mathcal{V}}$ is labeled by an 
index $\hat{\imath}$, e.g., $\hat{\imath}$ = $(i,j)$, with $\hat{\imath}(1)$ = $i$ and $\hat{\imath}(2)$ = $j$. 
Given the connectivity of the spatial grid defined by $\mathcal{E}_{s}$, we can for example identify $((i,j),(i,l))$ as an edge in $\hat{\mathcal{E}}$, but not $((i,j),(l,l))$.
Since $N$ = $2$ and $S$ = $3$, we have that $|\mathcal{M}|$ = $3^2$ = $9$. For each $\hat{\imath},\hat{\jmath} \in \hat{\mathcal{V}}$,
we can compute the transition probabilities in $\mathbf{Q} \in \mathbb{R}^{9 \times 9}$ from Equation \eqref{eqn:TransitionRatesProductState} as follows:  
\begin{eqnarray}
    q_{\hat{\imath} \hat{\jmath}} & = Pr\left(\mathbf{Y}_{k+1} = \hat{\jmath} ~|~ \mathbf{Y}_{k} = \hat{\imath} \right) = p_{\hat{\imath}(1)\hat{\jmath}(1)}p_{\hat{\imath}(2)\hat{\jmath}(2)}, \nonumber \\
    & \hspace{3cm} ~k \in \mathbb{Z}_+.
     \label{eqn:ProductSpaceTransitionProb}
\end{eqnarray}
 
We now prove that all robots will reach consensus on the feature distribution and it will converge to the reference distribution.
\begin{theorem}
Consider a group of $N$ robots, moving on a finite spatial grid with DTDS Markov chain dynamics \Cref{eqn:MarkovChain}, that update their opinions $f^a_k(s)$ for the feature distribution on the grid according to \Cref{eqn:feautePMF,eqn:Opinion_Update_eqn,eqn:FusedOccupancyVec,eqn:NextOccupancyVec}. Then for each robot $a \in \mathcal{A}$, $f^{a}_k(s)$ will converge to $f^{ref}(s)$ as $k \rightarrow \infty$ almost surely.
\end{theorem}
\begin{proof}
Suppose at initial time $k_0$, the locations of the robots on the spatial grid 
are given by the node $\hat{\imath} \in \hat{\mathcal{V}}$. Consider another set of robot locations at a future time $k_0 + k$, given by the node $\hat{\jmath} \in \hat{\mathcal{V}}$. The transition of the robots from configuration $\hat{\imath}$ to configuration $\hat{\jmath}$ in $k$ time steps corresponds to a random walk of length $k$ on the composite Markov chain $\mathbf{Y}_k$ from node $\hat{\imath}$ to node $\hat{\jmath}$. It also corresponds to a random walk by each robot $a$ on the spatial grid from node $\hat{\imath}(a)$ to node $\hat{\jmath}(a)$ in $k$ time steps.
By construction, the graph $\mathcal{G}_s$ is strongly connected and each of its nodes has a self-edge.
Thus, there exists a discrete time $n>0$ such that, for each robot $a$, there exists a random walk on the spatial grid from node $\hat{\imath}(a)$ to node $\hat{\jmath}(a)$ in $n$ time steps. Consequently, there always exists a random walk of length $n$ on the composite Markov chain $\mathbf{Y}_k$ from node $\hat{\imath}$ to node $\hat{\jmath}$, and 
therefore $\mathbf{Y}_k$ is an irreducible Markov chain. All states of an irreducible Markov chain belong to a single communication class. In this case, all states are \textit{positive recurrent};
as a result, each state of $\mathbf{Y}_k$ is visited infinitely often by the group of robots. 
Moreover, because the composite Markov chain is irreducible, we can conclude that $\cup_{k \in \mathbb{Z}_+} \mathcal{G}_c[k] = \mathcal{G}_0$, where $\mathcal{G}_0$ is the complete graph on the set of robots. Therefore $\mathcal{G}_0$ is strongly connected. Hence, each robot will meet every other robot at some node $s \in \mathcal{S}$ infinitely often. Since $\mathbf{Y}_{k}$ is irreducible and, from \Cref{eqn:NextOccupancyVec}, we have that $\theta^{a}_{k}(s) \leq \theta^{a}_{k+1}(s) \leq \theta^{ref}(s), ~\forall a \in \mathcal{A}, ~\forall s \in \mathcal{S}$, it follows from \Cref{eqn:FusedOccupancyVec,eqn:NextOccupancyVec}
that $\theta^{a}_{k}(s) \rightarrow \theta^{ref}(s)$ as $k \rightarrow \infty$. Consequently, $f^{a}_{k}(s) \rightarrow f^{ref}(s)$ as $k \rightarrow \infty$ almost surely.
\end{proof} 
\section{Numerical Simulation Results}\label{sec:MatSims}
In the numerical simulations, we consider a set of robots $\mathcal{A} = \{1,2,3,4\}$ moving on a 5.6 m $\times$ 5.6 m domain that is discretized into a square grid with $c = 8$ nodes on each side, with a distance $\delta = 0.7$ m between adjacent nodes.
The robots switch from one node to another at each time step according to the Markov chain dynamics in \Cref{eqn:MarkovChain}. The state transition probabilities $p_{ij}$, $i,j \in \mathcal{S} = \{1,2,\ldots,c^2\}$,  
that are associated with the spatial graph  $\mathcal{G}_{s}$ are computed from \Cref{eqn:TransitionMat_Elements}. We set the value of $\bar{l} = 0.8$ in \Cref{eqn:OccupancyGridFunction} in both the numerical and Software-In-The-Loop simulations. We distribute the features on the set of nodes 
$\mathcal{B}^{r} = \{ 19,20,21,26,30,34,38,42,46,51,52,53\}$, which represents a discrete approximation of a circular distribution on the grid. 
The set of neighbors $\mathcal{N}_k^a$ of robot $a$ at time $k$ consists of all robots that are located at the same node as robot $a$ at that time.

\begin{figure}[t]
    \centering
    \includegraphics[width=3.3in]{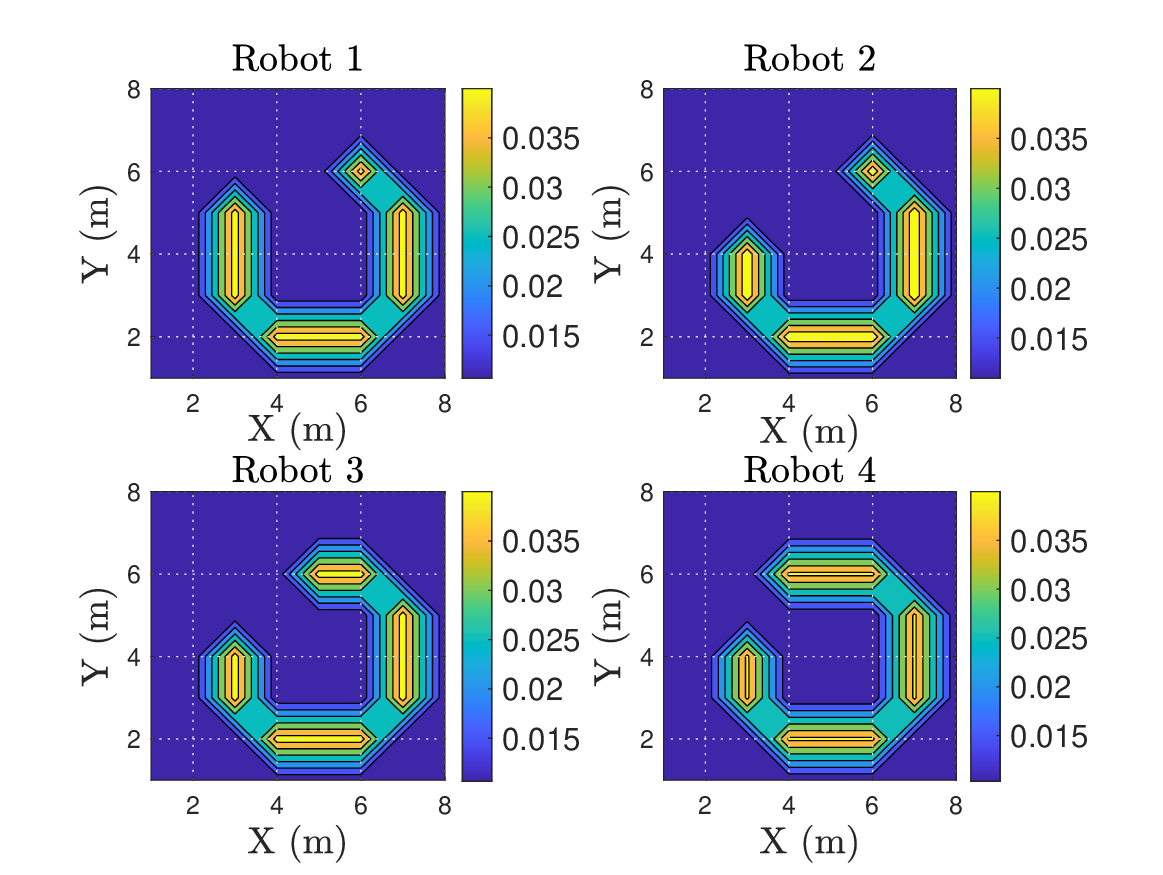}
    \caption{Feature PMF $f_k^a(s)$ of 4 robots at time $k=240$ s in the numerical simulation.}
    \label{fig:PMF_k_240}
\end{figure}

\begin{figure}[t]
    \centering
    \includegraphics[width=3.3in]{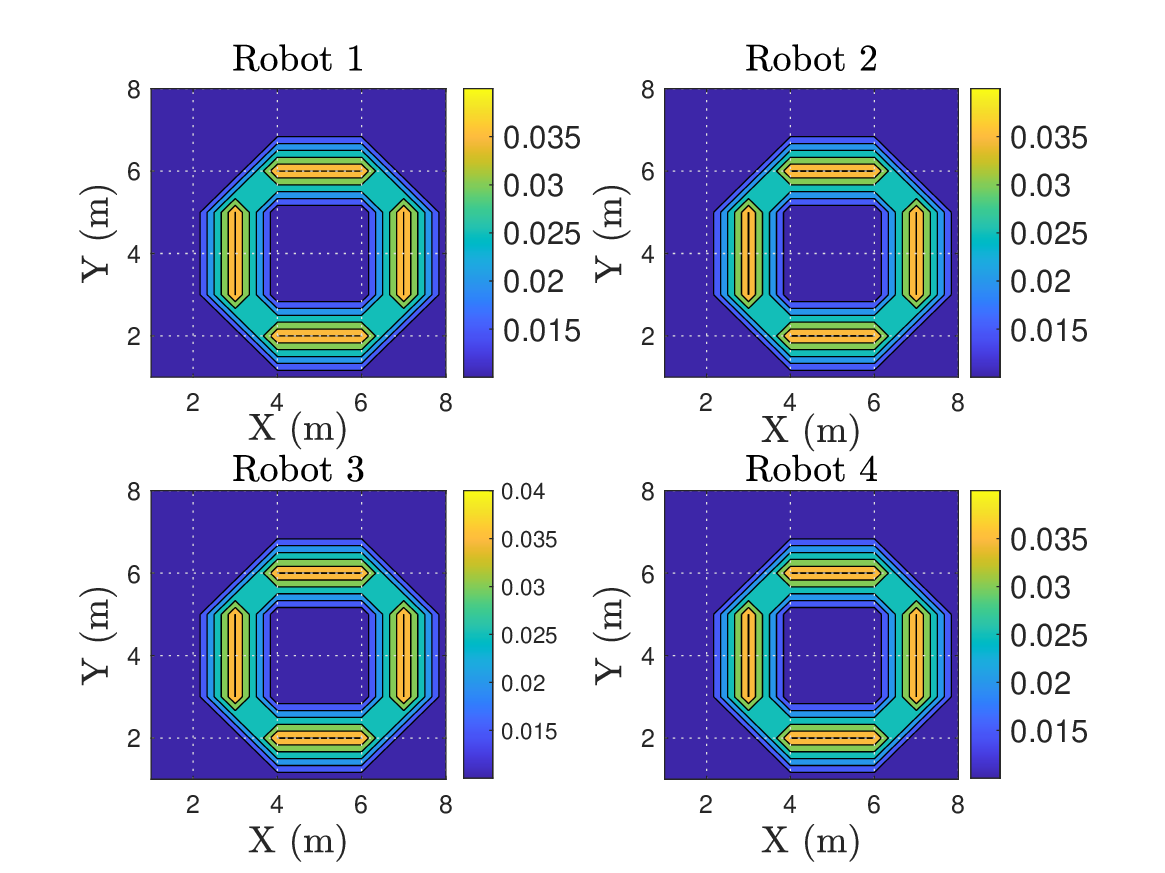}
    \caption{Feature PMF $f_k^a(s)$ of 4 robots at time $k=500$ s in the numerical simulation.}
    \label{fig:PMF_k_fin}
\end{figure}

\begin{figure}[t]
    \centering
    \includegraphics[width=3.3in]{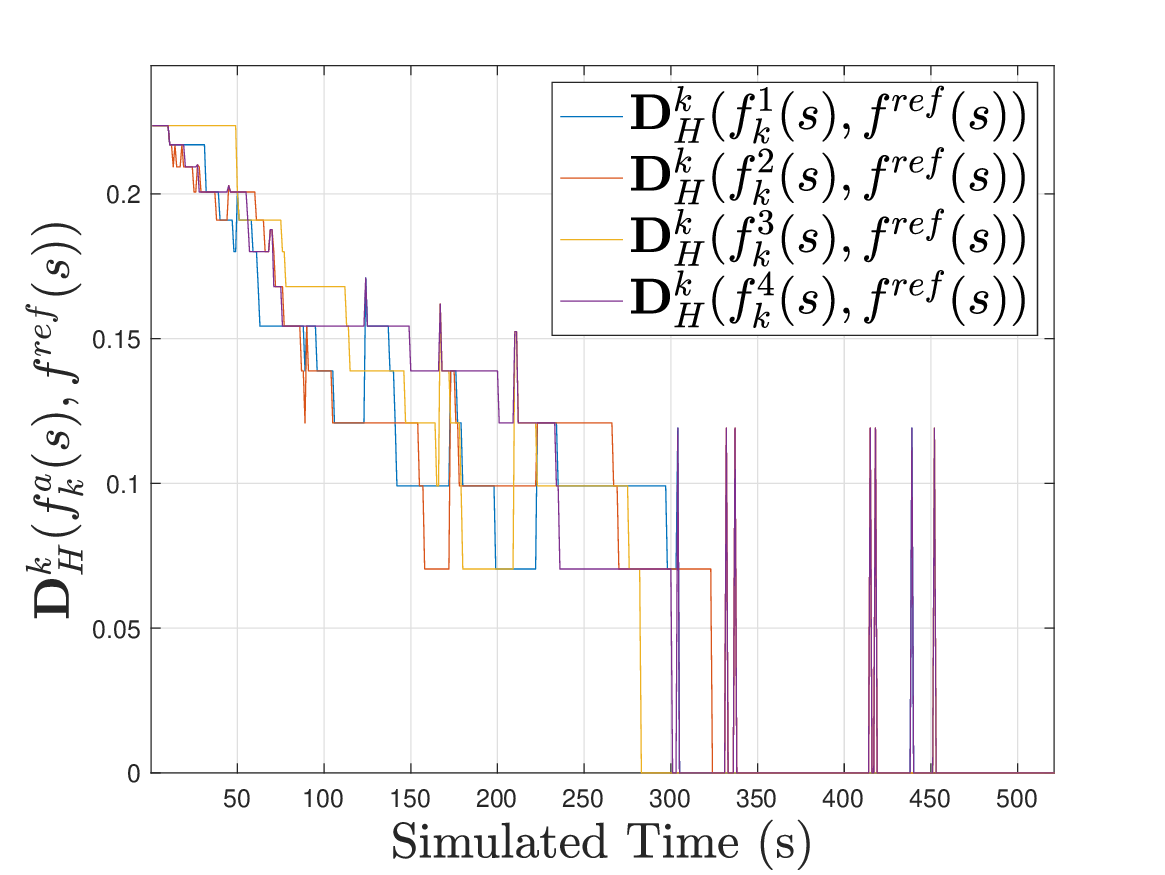}
    \caption{Time evolution of $\mathbf{D}_{H}(f^a_k(s),f^{ref}(s))$ for each robot $a \in \{1,2,3,4\}$ in the numerical simulation.}
    \label{fig:D_hell}
\end{figure}

All robots are initialized at uniformly random nodes in $\mathcal{S}$. Prior to exploration, the robots assume that all the grid nodes are unoccupied by features, and hence the vector of Hellinger distances is initially $\mathbf{D}_{H} = \mathbf{0} \in \mathbb{R}^{N \times 1}$. During their exploration of the grid, when robots encounter each other at the same node, they exchange their current feature PMFs and fuse them according to \Cref{eqn:Opinion_Update_eqn}. \Cref{fig:PMF_k_240} and \Cref{fig:PMF_k_fin} show the feature PMFs computed by each robot at $k=240$ s and $k=500$ s, respectively. We observe in \Cref{fig:PMF_k_240} that by $k=240$ s, all robots have partially reconstructed the feature PMF, with robot 4 having the closest reconstruction as measured by $\mathbf{D}_{H}(f^4_{240}(s),f^{ref}(s))\approx0.07$ in  \Cref{fig:D_hell}. 
In \Cref{fig:PMF_k_fin}, we see that all robots have successfully reconstructed the feature PMF by $k=500$ s. The robots' consensus on the reference PMF is also apparent from \Cref{fig:D_hell}, which shows that all Hellinger distances are zero at that time. Several distances $\mathbf{D}_{H}(f^{a}_{k}(s),f^{ref}(s))$ increase at times $k$ between 400 s and 500 s, due to the numerical inaccuracies in the fusion.

\begin{figure}
    \centering
    \includegraphics[width=3.3in]{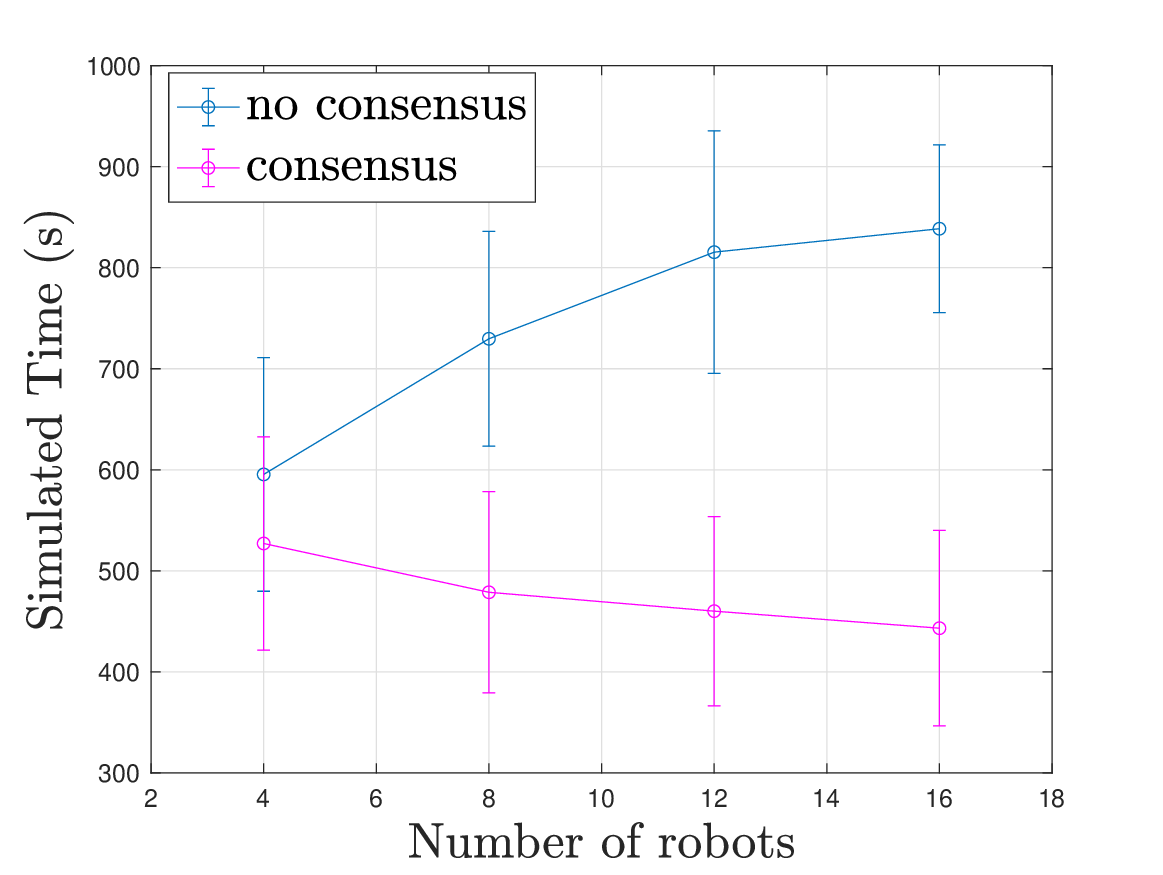}
    \caption{Time until convergence of robots' feature PMFs to the reference PMF in numerical simulations of the feature reconstruction strategy with and without consensus, for $N = \{4,8,12,16\}$ robots.}
    \label{fig:c_vs_nc_comp}
\end{figure}

We also ran Monte Carlo numerical simulations with different numbers of robots, $N = \{ 4,8,12,16 \}$, to investigate the effect of $N$ and the effect of consensus on the performance of the strategy in terms of the time for all robots' feature PMFs to converge to the reference PMF.  
\Cref{fig:c_vs_nc_comp} plots the resulting time, averaged over 100 simulations (error bars show standard deviations), until the feature PMFs of all $N$ robots converge to the reference PMF in the consensus and no-consensus cases.  
The figure shows that for $N=4$ robots, the mean time until convergence for both the consensus and no-consensus strategies are similar, with a significant overlap in their standard deviations. This indicates that for small numbers of robots, 
both strategies perform similarly. However, as $N$ increases, there is a widening gap between the mean times until convergence of the consensus and no-consensus strategies, with the times for the strategy with consensus being consistently lower. For $N=16$ robots, the strategy with consensus is faster than the one without consensus by a factor of  $\sim$2 (830 s/438 s). 

\section{Software-In-The-Loop Results}\label{sec:Expts}
We also validate our approach in Software-In-The-Loop (SITL) simulations using Gazebo and ROS Melodic. We simulate the same scenario as in the numerical simulations, with the set of robots $\mathcal{A}$ consisting of two  quadrotors, {\it Robot 1} and {\it Robot 2}. \Cref{fig:sitl_setup} illustrates the simulation setup. Video clips of the SITL simulations are included in the overview video \url{https://youtu.be/-Z4-DZrHwSM}.
The quadrotors are simulated using the Rotors \cite{furrer2016rotors} package in Gazebo, and the PX4 flight control stack is implemented in SITL to execute all low-level control tasks for each quadrotor. The discrete feature distribution that the quadrotors must reconstruct is represented by the ArUco markers on the ground plane, located at positions along the red dotted circle. 
The quadrotors fly at different altitudes (1 m and 2 m) to avoid collisions. This eliminates the need for obstacle detection and avoidance strategies, which are beyond the scope of this work. To detect the ArUco markers, each quadrotor is equipped with a simulated VGA resolution RGB camera that takes images at 30 fps and is oriented to face the ground plane. The image window size is heuristically adjusted to account for the difference in perspective resulting from the robots' difference in altitude. The quadrotors are assigned static IP addresses and exchange their feature distributions over wireless communication when they meet at the $x,y$ position of the same node (at different altitudes). We chose $T = 550$ s as the final time of the simulations. 

\subsection{System architecture}
\begin{figure}[t]
    \centering
    \includegraphics[width=3.4in]{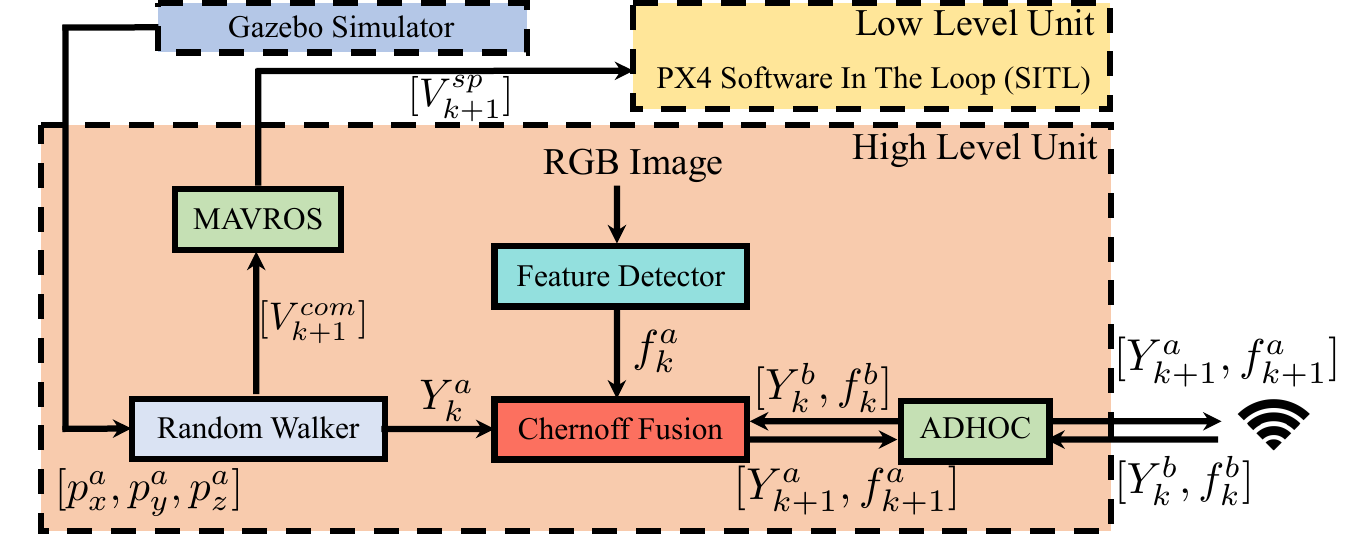}
    \caption{SITL simulation system architecture block diagram.} 
    \label{fig:sys_arch}
\end{figure}

A diagram of our system architecture is shown in \Cref{fig:sys_arch}. We use a hierarchical control scheme composed of low-level and high-level control blocks. 
The following is a description of each block in \Cref{fig:sys_arch}. The low-level controllers ({\it Low Level Unit}) use the ROS package MAVROS to generate the control commands for the simulated quadrotors from the high-level controller ({\it High Level Unit}). Gazebo provides global localization for the simulated quadrotors. It outputs the 3D position of each robot $a$, $[p_{x}^{a}, p_{y}^{a}, p_{z}^{a}]$, and its orientation.The 3D quadrotor positions from Gazebo are used to determine the quadrotors' current nodes on the grid and the nodes they should move to in the next time step, according to the DTDS Markov chain in \Cref{eqn:MarkovChain}. The new node locations are mapped to the commanded quadrotor velocities at time $k+1$, $V_{com}^{k+1}$, which are converted by {\it MAVROS} into velocity set points $V_{sp}^{k+1}$ and sent to the {\it Low Level Unit}. This block performs the corrections to the quadrotors' poses, which are rendered in Gazebo. An RGB image from the quadrotor's bottom-facing camera is used to detect an ArUco marker at the quadrotor's current node, and the feature PMF $f^a_k(s)$ is computed according to \Cref{eqn:feautePMF}. When two quadrotors are located at the same node, they exchange their feature PMFs using ZeroMQ \cite{hintjens2013zeromq}. \textit{Chernoff Fusion} block executes the fusion protocol in \Cref{algo:Chernoff_Fusion}. It computes the quadrotor's feature PMF from its own measurements and from the feature PMFs transmitted by other quadrotors at its current node through the {\it ADHOC} block.

\subsection{Simulation results}
\begin{figure}[t]
    \centering
    \includegraphics[width=3.3in]{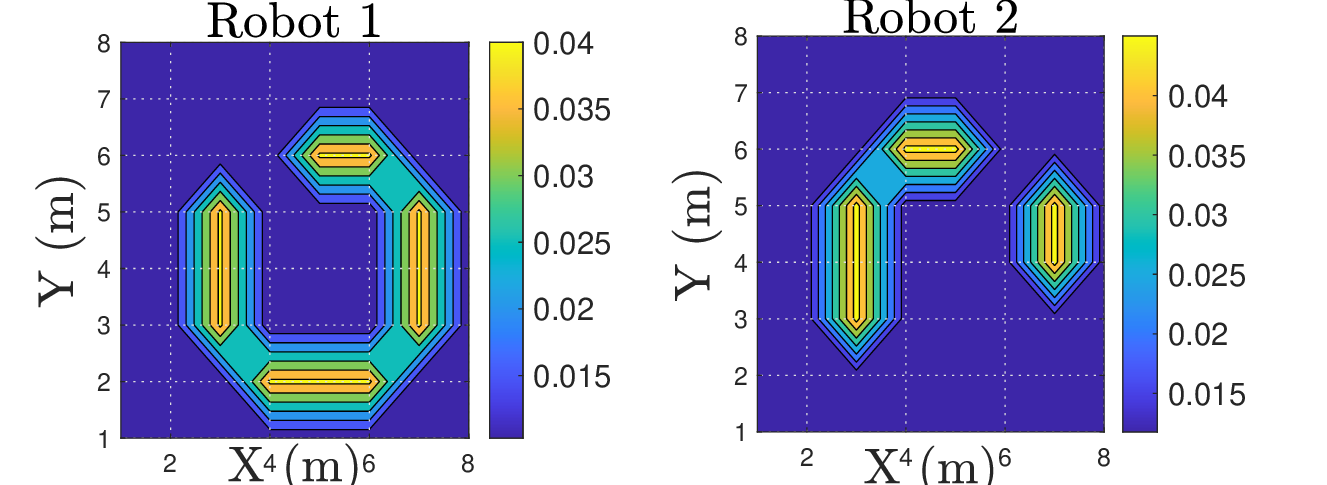}
    \caption{Feature PMF $f_k^a(s)$ of 2 robots at time $k=240$ s in the first run of the SITL simulation.}
    \label{fig:sil_dist_run1_t240}
\end{figure}

\begin{figure}[t]
    \centering
    \includegraphics[width=3.3in]{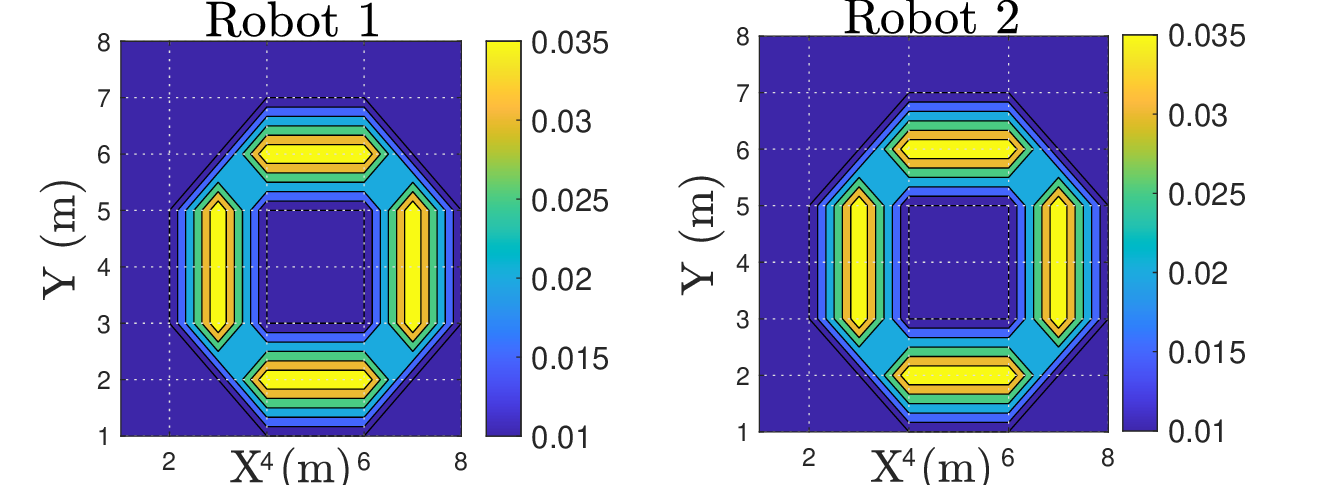}
    \caption{Feature PMF $f_k^a(s)$ of 2 robots at time $k=330$ s in the first run of the SITL simulation.}
    \label{fig:sil_dist_run1_t330}
\end{figure}

\begin{figure}[t]
    \centering
    \includegraphics[width=3.3in]{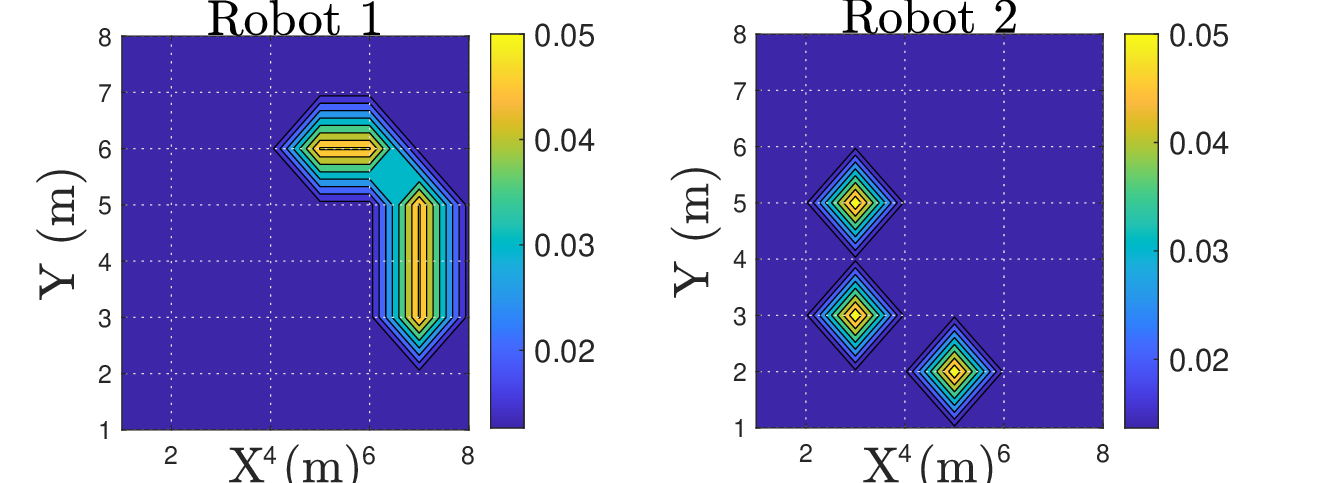}
    \caption{Feature PMF $f_k^a(s)$ of 2 robots at time $k=240$ s in the second run of the SITL simulation.}
    \label{fig:sil_dist_run2_t240}
\end{figure}

\begin{figure}[h]
    \centering
    \includegraphics[width=3.3in]{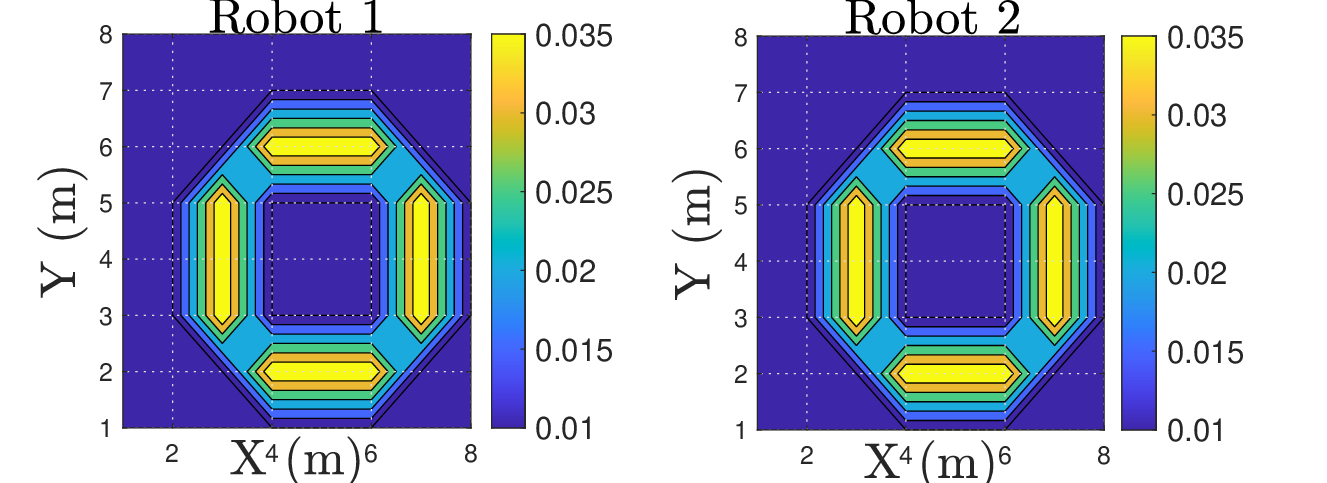}
    \caption{Feature PMF $f_k^a(s)$ of 2 robots at time $k=530$ s in the second run of the SITL simulation.}
    \label{fig:sil_dist_run2_t530}
\end{figure}

\Cref{fig:sil_dist_run1_t240}-\Cref{fig:sil_dist_run2_t530} plot the feature PMFs computed by both quadrotors during two SITL simulation runs. \Cref{fig:sil_dist_run1_t240} and \Cref{fig:sil_dist_run1_t330} show the feature PMFs at $k=240$ s ($4$ min) and $k=330$ s ($5.5$ min), respectively, of the first simulation run. The figures indicate that {\it Robot 1} reconstructs most of the feature distribution within $4$ min, and both robots fully reconstruct the distribution within $5.5$ min. 

\Cref{fig:sil_dist_run2_t240} and \Cref{fig:sil_dist_run2_t530} plot the feature PMFs at $k=240$ s ($4$ min) and $k=530$ s ($\sim8.8$ min), respectively, of the second simulation run. \Cref{fig:sil_dist_run2_t240} shows that in this simulation run, the two robots do not meet and exchange feature PMFs within the first $4$ min, since their feature PMFs are completely distinct at that time. By $\sim8.8$ min, \Cref{fig:sil_dist_run2_t530} shows that both robots have fully reconstructed the feature distribution, which matches the reconstructed distribution in  \Cref{fig:sil_dist_run1_t330}. Thus, \Cref{fig:sil_dist_run1_t240} through  \Cref{fig:sil_dist_run2_t530} demonstrate that our approach ultimately results in accurate feature reconstruction, but that the convergence time to full reconstruction can differ between runs due to the randomness in the robot paths over the grid. 

\subsection{Considerations for real-world implementation}\label{subsec:realimpl}
The performance of our approach in real-world environments will be affected by aerodynamic interactions between quadrotors, uncertainty in positioning, and feature occlusion by a quadrotor that enters another's field of view. In the SITL simulations, it is difficult to accurately simulate the aerodynamic disturbance on a quadrotor caused by the downwash of a quadrotor above it. These disturbances can be rejected by incorporating a robust disturbance observer \cite{mishra2019robust, mishradob} into the quadrotor's low-level flight control strategy. Alternatively, the quadrotors could fly at the same altitude and employ controllers for inter-robot collision avoidance, e.g. using control barrier functions \cite{wang2017safe}. Quadrotors may miss feature detections due to misalignment of their camera image window with the ground or to occlusion of the feature by another quadrotor flying below it. Such misalignments and occlusions sometimes occurred in our SITL simulations; however, the accuracy of the feature reconstruction despite these occurrences demonstrates the robustness of our approach to the resulting missed feature detections.

\section{Conclusion and Future Work}\label{sec:Conc}
In this work, we propose a decentralized multi-robot strategy for reconstructing a discrete feature distribution on a finite spatial grid. Robots update their estimate of the feature distribution using their own measurements during random-walk exploration and estimates from nearby robots, combined using a distributed Chernoff fusion protocol. Our strategy extends established results on consensus of opinion pools for fixed, strongly connected networks to networks with Markovian switching dynamics. We provide theoretical guarantees on convergence to  the ground truth feature distribution in an almost sure sense, and we validate the strategy in both numerical simulations and SITL simulations with quadrotors.

We note that our strategy is agnostic to the source of the information used to reconstruct the feature distribution; it values the information gained from exploration and other robots equally. This can result in suboptimal convergence rates to consensus on the ground truth distribution, potentially exceeding the operational flight times of small aerial robots with limited battery life. To increase the convergence rate to consensus, we propose to modify the robots' exploration strategy from unbiased random walks to random walks that are biased in directions that increase information gain from individual robots' on-board sensor measurements and also ensure frequent encounters between robots. \revfin{Another possible way to decrease the time to consensus is by relaxing the constraint $r_{comm} < 0.5\delta$, allowing robots to communicate with robots at other nodes. This would require an analysis of whether consensus is still ensured 
by reformulating the composite Markov chain and determining whether it is irreducible and positive recurrent.
In addition, while P\'{o}lya's recurrence theorem \cite{polya1921aufgabe} guarantees that our results on irreducibility and positive recurrence do not extend 
to random walks on infinite 3D lattices, it would be interesting to investigate  whether they are valid for 
finite 3D grids.}
\revfin{Finally, it would be useful to derive an analytical formulation of the expected time until consensus, if possible, which would provide a more rigorous basis for selecting the final time $T$.}

\bibliographystyle{unsrt}
\bibliography{main}

\begin{thebibliography}{10}

\bibitem{burgard2005coordinated}
Wolfram Burgard~\textit{et al}.
\newblock Coordinated multi-robot exploration.
\newblock {\em IEEE Transactions on Robotics}, 21(3):376--386, 2005.

\bibitem{nagatani2013emergency}
Keiji Nagatani~\textit{et al}.
\newblock Emergency response to the nuclear accident at the {Fukushima Daiichi
  Nuclear Power Plants} using mobile rescue robots.
\newblock {\em Journal of Field Robotics}, 30(1):44--63, 2013.

\bibitem{mendoncca2016cooperative}
Ricardo Mendonça~\textit{et al}.
\newblock A cooperative multi-robot team for the surveillance of shipwreck
  survivors at sea.
\newblock In {\em OCEANS 2016 MTS/IEEE Monterey}, pages 1--6, 2016.

\bibitem{mishra2021tmech}
Shatadal Mishra~\textit{et al}.
\newblock Autonomous vision-guided object collection from water surfaces with a
  customized multirotor.
\newblock {\em IEEE/ASME Transactions on Mechatronics}, 26(4):1914--1922, 2021.

\bibitem{sukkar2019multi}
Fouad Sukkar~\textit{et al}.
\newblock Multi-robot region-of-interest reconstruction with {D}ec-{MCTS}.
\newblock In {\em 2019 International conference on robotics and automation
  (ICRA)}, pages 9101--9107. IEEE, 2019.

\bibitem{mahdoui2018cooperative}
Nesrine Mahdoui~\textit{et al}.
\newblock Cooperative frontier-based exploration strategy for multi-robot
  system.
\newblock In {\em 2018 13th Annual Conference on System of Systems Engineering
  (SoSE)}, pages 203--210. IEEE, 2018.

\bibitem{furrer2016rotors}
Fadri Furrer~\textit{et al}.
\newblock {\em RotorS---A Modular Gazebo MAV Simulator Framework}, pages
  595--625.
\newblock Springer International Publishing, Cham, 2016.

\bibitem{howard2006experiments}
Andrew Howard~\textit{et al}.
\newblock Experiments with a large heterogeneous mobile robot team:
  Exploration, mapping, deployment and detection.
\newblock {\em The International Journal of Robotics Research},
  25(5-6):431--447, 2006.

\bibitem{husain2013mapping}
Ammar Husain~\textit{et al}.
\newblock Mapping planetary caves with an autonomous, heterogeneous robot team.
\newblock In {\em IEEE Aerospace Conference}, pages 1--13, 2013.

\bibitem{li2019fully}
Xianwei Li~\textit{et al}.
\newblock Fully distributed consensus control for linear multiagent systems: A
  reduced-order adaptive feedback approach.
\newblock {\em IEEE Transactions on Control of Network Systems}, 7(2):967--976,
  2020.

\bibitem{wei2017consensus}
Baishen Wei~\textit{et al}.
\newblock Consensus labeled multi-{Bernoulli} filtering for distributed space
  debris tracking.
\newblock In {\em International Conference on Control, Automation and
  Information Sciences (ICCAIS)}, pages 203--208, 2017.

\bibitem{xia2015structural}
Weiguo Xia~\textit{et al}.
\newblock Structural balance and opinion separation in trust–mistrust social
  networks.
\newblock {\em IEEE Transactions on Control of Network Systems}, 3(1):46--56,
  2016.

\bibitem{ren2004consensus}
Wei Ren and R.W. Beard.
\newblock Consensus of information under dynamically changing interaction
  topologies.
\newblock In {\em American Control Conference (ACC)}, volume~6, pages
  4939--4944, 2004.

\bibitem{mesbahi2010graph}
Mehran Mesbahi and Magnus Egerstedt.
\newblock {\em Graph theoretic methods in multiagent systems}.
\newblock Princeton University, Princeton, NJ, 2010.

\bibitem{olfati2004consensus}
Reza Olfati-Saber and Richard~M. Murray.
\newblock Consensus problems in networks of agents with switching topology and
  time-delays.
\newblock {\em IEEE Transactions on Automatic Control}, 49(9):1520--1533, 2004.

\bibitem{kegeleirs2019random}
Miquel Kegeleirs~\textit{et al}.
\newblock Random walk exploration for swarm mapping.
\newblock In {\em Towards Autonomous Robotic Systems}, pages 211--222, Cham,
  2019. Springer International Publishing.

\bibitem{shirsat2020multirobot}
Aniket Shirsat~\textit{et al}.
\newblock Multi-robot target search using probabilistic consensus on discrete
  {Markov} chains.
\newblock In {\em IEEE International Symposium on Safety, Security, and Rescue
  Robotics (SSRR)}, pages 108--115, 2020.

\bibitem{shirsat2021decentralized}
Aniket Shirsat and Spring Berman.
\newblock Decentralized multi-target tracking with multiple quadrotors using a
  {PHD} filter.
\newblock In {\em AIAA Scitech 2021 Forum}, 2021.

\bibitem{bailey2012conservative}
Tim Bailey~\textit{et al}.
\newblock On conservative fusion of information with unknown {non-Gaussian}
  dependence.
\newblock In {\em 15th International Conference on Information Fusion}, pages
  1876--1883, 2012.

\bibitem{battistelli2014kullback}
Giorgio Battistelli and Luigi Chisci.
\newblock {Kullback–Leibler} average, consensus on probability densities, and
  distributed state estimation with guaranteed stability.
\newblock {\em Automatica}, 50(3):707--718, 2014.

\bibitem{degroot1974reaching}
Morris~H. Degroot.
\newblock Reaching a consensus.
\newblock {\em Journal of the American Statistical Association},
  69(345):118--121, 1974.

\bibitem{calafiore2009distributed}
Giuseppe~C. Calafiore and Fabrizio Abrate.
\newblock Distributed linear estimation over sensor networks.
\newblock {\em International Journal of Control}, 82(5):868--882, 2009.

\bibitem{farrell2009generalized}
William~J. Farrell and Chidambar Ganesh.
\newblock Generalized {Chernoff} fusion approximation for practical distributed
  data fusion.
\newblock In {\em International Conference on Information Fusion}, pages
  555--562, 2009.

\bibitem{grimmett2001probability}
Geoffrey Grimmett and David Stirzaker.
\newblock {\em Probability and random processes}.
\newblock Oxford University Press, 2001.

\bibitem{horn1990matrix}
Roger~A. Horn and Charles~R. Johnson.
\newblock {\em Matrix analysis}.
\newblock Cambridge University Press, 1990.

\bibitem{levin2017markov}
David~A. Levin and Yuval Peres.
\newblock {\em Markov chains and mixing times}, volume 107.
\newblock American Mathematical Society, 2017.

\bibitem{hintjens2013zeromq}
Pieter Hintjens.
\newblock {\em ZeroMQ: messaging for many applications}.
\newblock O'Reilly Media, Inc., 2013.

\bibitem{mishra2019robust}
Shatadal Mishra~\textit{et al}.
\newblock {Robust attitude control for quadrotors based on parameter
  optimization of a nonlinear disturbance observer}.
\newblock {\em Journal of Dynamic Systems, Measurement, and Control}, 141(8),
  03 2019.
\newblock 081003.

\bibitem{mishradob}
Shatadal Mishra and Wenlong Zhang.
\newblock A disturbance observer approach with online {Q}-filter tuning for
  position control of quadcopters.
\newblock In {\em American Control Conference (ACC)}, pages 3593--3598, 2017.

\bibitem{wang2017safe}
Li~Wang~\textit{et al}.
\newblock Safe certificate-based maneuvers for teams of quadrotors using
  differential flatness.
\newblock In {\em IEEE International Conference on Robotics and Automation
  (ICRA)}, pages 3293--3298, 2017.

\bibitem{polya1921aufgabe}
Georg P{\'o}lya.
\newblock {\"U}ber eine aufgabe der wahrscheinlichkeitsrechnung betreffend die
  irrfahrt im stra{\ss}ennetz.
\newblock {\em Mathematische Annalen}, 84(1):149--160, 1921.

\end{thebibliography}
\end{document}